# Identification of mental fatigue in language comprehension tasks based on EEG and deep learning

Chunhua Ye [b], Zhong Yin [a, b, *], Chenxi Wu [b], Xiayidai Abulaiti [b],

Yixing Zhang [b], Zhenqi Sun [b], and Jianhua Zhang [c]

a. *Engineering Research Center of Optical Instrument and System, Ministry of Education, Shanghai Key Lab of Modern Optical System, University of Shanghai for Science and Technology, Shanghai, 200093, PR China*

b. *School of Optical-Electrical and Computer Engineering, University of Shanghai for Science and Technology, Shanghai, 200093, PR China*

c. *OsloMet Artificial Intelligence Lab, Department of Computer Science, Oslo Metropolitan University, Oslo, N-0130, Norway*

**Abstract**: Mental fatigue increases the risk of operator error in language comprehension tasks. In order to prevent operator performance degradation, we used EEG signals to assess the mental fatigue of operators in human-computer systems. This study presents an experimental design for fatigue detection in language comprehension tasks. We obtained EEG signals from a 14-channel wireless EEG detector in 15 healthy participants. Each participant was given a cognitive test of a language comprehension task, in the form of multiple choice questions, in which pronoun references were selected between nominal and surrogate sentences.

In this paper, the 2400 EEG fragments collected are divided into three data sets according to different utilization rates, namely 1200s data set with 50% utilization rate, 1500s data set with 62.5% utilization rate, and 1800s data set with 75% utilization rate. In the aspect of feature extraction, different EEG features were extracted, including time domain features, frequency domain features and entropy features, and the effects of different features and feature combinations on classification accuracy were explored. In terms of classification, we introduced the Convolutional Neural Network (CNN) method as the preferred method, It was compared with Least Squares Support Vector Machines(LSSVM),Support Vector Machines(SVM),Logistic Regression (LR), Random Forest(RF), Naive Bayes (NB), K-Nearest Neighbor (KNN) and Decision Tree(DT).According to the results, the classification accuracy of convolutional neural network (CNN) is higher than that of other classification methods. The classification results show that the classification accuracy of 1200S dataset is higher than the other two datasets. The combination of Frequency and entropy feature and CNN has the highest classification accuracy, which is 85.34%.

---
* Corresponding author: Zhong Yin, Tel.: +86 21 55271064. E-mail address: yinzhong@usst.edu.cn.
Address: Jungong Road 516, Yangpu District, Shanghai 200093, P. R. China.

## 1. Introduction

In the human-computer interaction environment, the operator's functional state affects the task performance [1].One challenge to maintaining task performance is that cognitive work leads to mental fatigue, which is characterized by reduced mental alertness and reduced performance [2].In a modern world of increased work stress, extended working hours, poor sleep quality or night shifts for operators in the human-computer interaction industry, mental fatigue is a common problem in today's workplace, which can interfere with productivity and overall cognitive function. According to the research, in the fields of public transportation[3], manufacturing industry[4] and nuclear power plant[5], human operators are an important factor leading to serious accidents [6].Performance degrades over time, which is also known as the effect of time on tasks. Mental fatigue has a negative impact on many cognitive functions. As the operation progresses, fatigue accumulates, which greatly reduces the operator's ability to understand, react and quickly solve problems in the operation, and will lead to wrong processing tasks [7].

To reduce risk, assessing the operator's functional status becomes a key factor. The operator's functional state can be evaluated from three main aspects: situational awareness, mental load and mental fatigue [8]. Mental fatigue is one of the key factors affecting human operational errors. Traditional measurement methods of mental fatigue can be divided into two categories: subjective measurement and objective measurement [9].The subjective mental fatigue measurement method requires the subjects to subjectively evaluate their mental fatigue degree, which can be evaluated by means of questionnaire, etc., while the objective method evaluates mental fatigue by quantifying the performance of the subjects on specific tasks, such as some behavioral data [10].

Traditional measurement methods have their limitations. According to research, physiological signals are considered to be the most reliable method to assess the operator's mental fatigue state, because physiological signals start to change before any external signs of mental fatigue appear [11]. In order to solve the problem caused by mental fatigue in human-computer interaction task, the automatic recognition of fatigue state is very important. Researchers have used a variety of physiological signals to study fatigue [12], including respiration, electrocardiogram (ECG), electromyogram (EMG), electroophthalmogram (EOG), and electroencephalogram (EEG) [13]-[15].Among these physiological signals, EEG can directly reflect the neural activity of the brain and is one of the most popular methods to detect fatigue [16].Because EEG signals are time-varying signals, there are too many variables [17], such as different brain shapes among individuals, different EEG forms, and excessive EEG interference on the scalp. Therefore, it is particularly important to preprocess and extract features of EEG signals.

In order to ensure that the automatic classification of EEG signals can achieve a high classification accuracy, machine learn-based methods have attracted extensive attention [12].The purpose of this study is to explore the effective features and feature combinations for identifying mental fatigue in EEG, and to explore a suitable classification method by using traditional machine learning methods and deep learning methods to conduct classification tests on features. Traditional machine learning has certain advantages in processing low-dimensional data. However, the high-dimensional nature of EEG features may reduce the stability of EEG recognition rate training data-driven model [18].To solve this problem, we introduce a convolutional neural network to identify useful mesoencephalograms to indicate the characteristics of distinguishable cognitive states. CNN not only can automatically extract features, but also has obvious advantages in processing high-dimensional data.

In this paper, an experiment of language comprehension task is designed to detect the mental fatigue of operators in human-computer interaction. The data set is divided into three data sets according to different utilization. For each EEG segment, the extracted features include frequency domain features, time domain features, and entropy features. By using nine classification methods, different feature combinations are explored, and it is found that in the same classifier, the classification accuracy of frequency domain feature and entropy feature combination is slightly improved. The input of frequency domain and entropy characteristics in the convolutional neural network is the best classification method in this study. This study also found that the lower the data utilization rate in the same classification method, the higher the classification accuracy.

## 2. Related works

Electroencephalography (EEG) is a medical imaging technique [19] that measures the electrical activity of the scalp produced by the brain, that is, the electrical fluctuations of the surface of the scalp caused by the ionic current within the mental state and recorded in chronological order. The signals collected by EEG are transformed into frequency domain by Discrete Fourier Transform (DFT), Fast Fourier Transform (FFT) and other algorithms, and include continuous spectral signals[20]. These signals are divided into five frequency bands according to frequency :Delta (1-4Hz), Theta (4-7Hz), Alpha (8-13Hz), Beta (13-30Hz), Gamma (30-40Hz).In [21], Klimesch found that EEG oscillations in the alpha and theta bands specifically reflected cognitive and memory performance. When subjects were asked to stay awake while sleepy, the lower alpha power increased. If the subjects were allowed to sleep, alpha power went down and theta power went up. According to a. Craig's results in [11], when A person is tired, slow-wave activity in the whole cortex, theta and 1 and 2 bands increases, while delta wave activity does not change significantly. The brain loses capacity and slows its activity, and trying to stay alert can lead to increased beta activity.

In order to detect the mental fatigue state of operators, the most popular method is to use machine learning method to identify mental fatigue from EEG signals [22].The usual approach is to extract a set of features that are used to train a classifier to maximize the distinction between fatigue and normal EEG signals. Li extracted three frequency band features, (4-8 Hz), (8-13 Hz) and (13-30 Hz), and used BN, SVM,LR, KNN, and RF to classify depression and normal, and achieved good classification results [23].The entropy feature extraction method can be used to detect fatigue well.Hu extracted four entropy features of sample entropy, fuzzy entropy, approximate entropy and spectral entropy, and compared them with traditional machine learning classification methods. The results showed that RF classifier had the best accuracy [24]. Polat & Güneş using decision tree classifier and fast Fourier transform of the hybrid system to detect EEG signals of seizures [25]. Setiawan extracted the power spectral density characteristics, introduced the correlation vector machine method as the priority classification method, and compared with SVM, the results were better [26].By analyzing the frequency distribution of power spectral density (PSD), Ozmen proposed an improved recognition rate method, and finally obtained an accuracy rate of 83.06% for dichotomies and 91.85% for multiple classifications [27].

The classification of EEG signals using deep learning technology is a popular direction. It can not only learn the cascade representation directly from the original EEG data, but also further purify the extracted features[28], calculate the output value through forward propagation, and train the classifier through back propagation to adjust the weight and bias. Yang proposed a frequency complementary feature map selection (FCMS) scheme, and combined CNN features with enhanced CSP features to improve the accuracy of classification [29].Gao used the combination of traditional methods and deep learning technology to propose a convolutional neural network (RN-CNN) method based on repetition to detect fatigue driving [30].Jana et al filtered the original EEG signal data and generated spectral feature matrix, and finally used one-dimensional convolutional network for seizure detection [31].Gao used CNN to detect driver fatigue and obtained a good classification effect [32].

## 3. Materials and Methods

## 3.1 Collection of experimental data

EEG signals were collected from 15 healthy subjects using a wireless head-worn EEG detector. All tasks were selected from graduate students who had signed informed consent prior to the experiment. Before the formal test, each participant received 10 language comprehension task attempts with the same task conditions. During the test, five isochronous experimental modules were conducted. Second stimulus: display generation of words; Third stimulus: choose pronoun reference. Display time of stimulus events (as shown in Figure 1) : each stimulus event will be displayed for 3000 ms before entering the next stimulus event; after the end of the second stimulus, the white screen will last for 2500ms and the cross will be displayed for 500ms before entering the next cycle. Note: The response time was 5500 ms (including 3000 ms third stimulus and 2500 ms white screen).Every two modules are spaced 2 minutes apart. The experiment has 5 modules, each module contains 40 sentences, a total of 200 sentences (200=5*40).In the experiment, three stimulus events are displayed sequentially. It takes 12s for the subject to complete a task, and 2400s to complete the whole experiment.

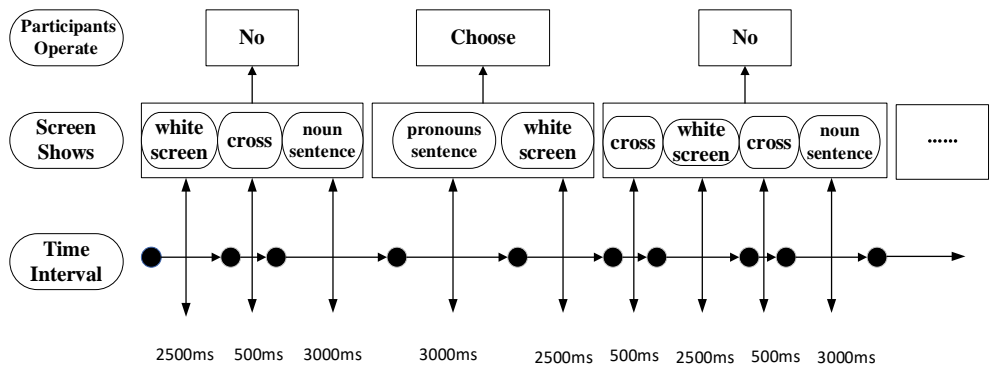

Fig. 1. Language understanding of experimental procedures

## 3.2 Preprocessing and feature extraction of experimental data

According to the international 10-20 system, 14 channels are selected for further analysis, namely AF3, AF4, F3, F4, P7, P8, O1, O2, F7, F8, FC5, FC6, T7 and T8.Sampled at 128Hz, the collected EEG signals were preprocessed through a linear finite impulse response (FIR) filter and independent variable analysis, and 4-45Hz bandpass filtering was performed. According to the process of the experiment, the useful experimental data were divided according to the interval of one experiment (12s), and the EEG data during the interval was removed. To facilitate processing, one experiment was equally divided into 12 one-second experiments with a total of 2,400 EEG fragments. For the part of the experiment in which the experimental subjects did not make a choice, the average of other subjects' response time to the experiment was used as the reaction time of the experimental subjects.

Common feature extraction methods include power spectral density (PSD), statistics and entropy measure[33]-[34]. The PSD of a time series describes the power distribution in the signal as a function of frequency. The Fourier transform is used to transform the EEG signal from the time domain to the frequency domain before calculating the PSD. For each EEG segment of all channels, there are four PSD characteristics, namely, average power in the bands of Theta (4-8 Hz), Alpha (8-13 Hz), Beta (14-30 Hz), and Gamma (31-40 Hz).Each EEG segment extracted 56 features in the frequency domain. Due to the non-stationarity of the data itself, its performance is poor. The EEG signal is divided into shorter segments, and the signal of each segment is assumed to be stable. Then statistical analysis was performed on EEG signals to calculate five parameters, namely mean value, variance, zero-crossing Rate (ZCR), Kurtosis, Skewness. Each EEG segment extracted 70 time-domain features. Entropy measure method is robust to evaluate the regularity and predictability of complex systems. The Entropy measures extracted are Shannon Entropy (SE) and SpecEn. Each EEG segment extracted 28 entropy features.

### 3.3 CNN as mental fatigue classifier

*3.3.1 CNN Input Construction*

The EEG interface system uses the EMOTIV Epoc+ wireless portable encephalograph device to collect EEG signals. The 10-20 system is an internationally recognized method for describing and applying the location of scalp electrodes and subcortical areas of the brain[35]. In the upper left corner of Figure 2, 14 EEG electrodes selected from the international 10-20 system are used as test points for this subject's data set. In the electroencephalogram (EEG) spectrum, each electrode is physically adjacent to several electrodes that record electrical signals in a particular area of the brain. In order to maintain spatial information between multiple adjacent channels, one-dimensional features were first transformed into two-dimensional plane features according to the electrode distribution diagram. Frequency domain features, time domain features and entropy features were transformed into two-dimensional plane features of 67*67 by topoplot respectively, and each two-dimensional plane feature was transformed into 34*34 dimensions by down-sampling (As shown in Figure 2).

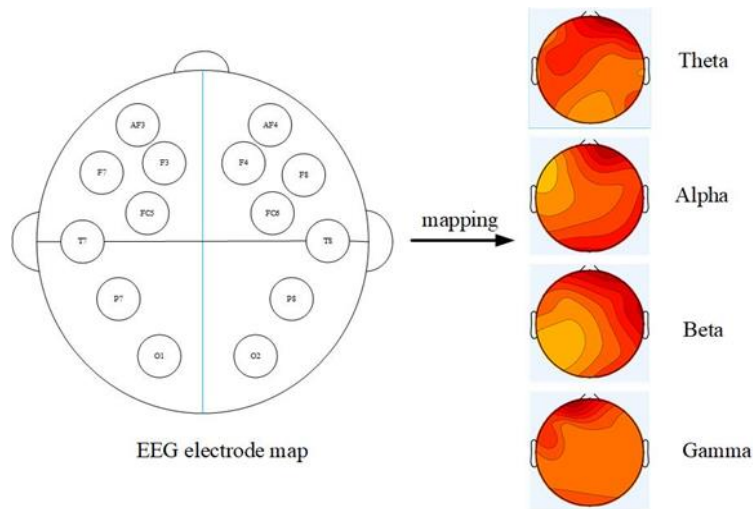

Fig. 2. The EEG electrode map was drawn into N 34*34 dimensional two-dimensional EEG topographical maps.

We obtained that each EEG fragment has four two-dimensional planes in frequency domain, five two-dimensional planes in time domain, and two two-dimensional planes in entropy. We then stacked these planes into 3D EEG cubes and used them as input to CNN (Figure 3).In order to further explore the influence of features and feature combinations on classification accuracy, input feature combinations are carried out, as shown in Table 1.

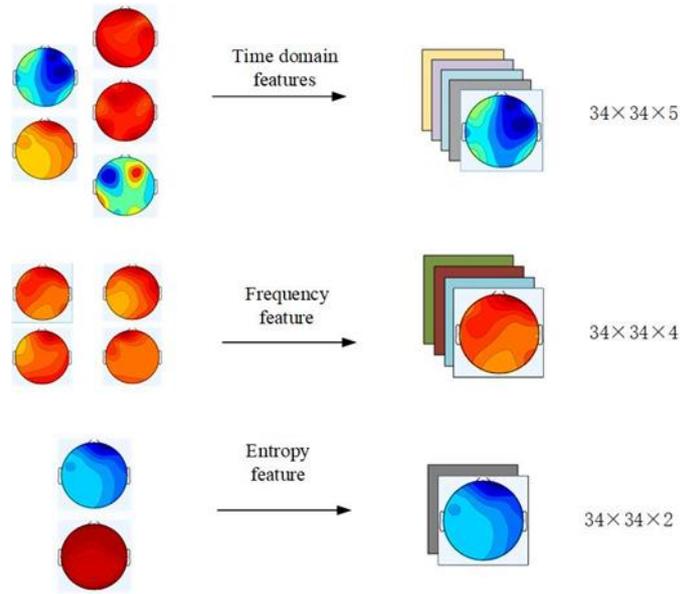

Fig. 3. Stack 2D EEG topographic maps into 3D EEG cubes

Table 1. 3D EEG cube input format

| Input features | Input format |
| --- | --- |
| Frequency features | 34×34×4 |
| Time domain feature | 34×34×5 |
| Entropy feature | 34×34×2 |
| Frequency and time domain feature | 34×34×9 |
| Frequency and entropy feature | 34×34×6 |
| Entropy and time domain feature | 34×34×7 |
| Frequency and time domain and entropy feature | 34×34×11 |

### 3.3.2 CNN model

In essence, a convolutional network is an input-output mapping[36]. It can learn a large number of mapping relations between inputs and outputs without any precise mathematical expression between inputs and outputs. As long as the convolutional network is trained with known patterns, the network has the mapping ability between input and output pairs. CNN is a feedforward neural network, which can extract its topology structure from a two-dimensional image, and adopt back propagation algorithm to optimize the network structure and solve unknown parameters in the network. In this paper, a typical convolutional neural network is used to extract the characteristics of the input, as shown in Figure 4.

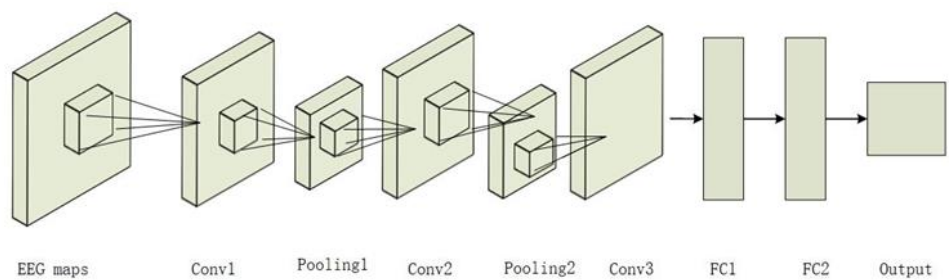

Fig. 4. The structure of the CNN

The Conv1 layer is a convolutional layer, and an important feature of the convolution operation is that the Conv1 layer enhances the original signal characteristics and reduces the noise[37]. It is composed of four filters, each of which has a size of 3*3.We select padding= 'same' to fill, the filled value is calculated by the algorithm based on the convolution kernel size, in order to make the output size equal to the input. The input to the Pooling1 layer is then calculated using the Relu activation function.

The Relu formula is as follows: $F(x)= Max (0,x)$. If the input x is less than 0, the output is equal to 0;If the input x is greater than 0, then the output is equal to the input. When x is greater than 0, the gradient is constant as 1, there is no gradient dissipation problem, and the convergence is fast. It increases the sparsity of the network. When x is less than 0, the output of this layer is 0. The more neurons that are 0 after training, the greater the sparsity, the more representative the extracted features will be and the stronger the generalization ability will be. In other words, if the same effect is obtained, the less neurons are actually active, the better the generalization performance of the network is.

Pooling1 layer is a lower sampling layer[38]. Max Pooling is used. Sub-sampling is conducted for images based on the principle of local correlation, which can reduce data processing capacity and retain useful information. Each cell in the eigengraph is connected to a 2*2 neighborhood of the Conv1 corresponding eigengraph.Pooling1 layer adds up the four inputs to each cell, multiplied by a trainable parameter plus a trainable bias. The 2*2 receptive fields of each cell do not overlap, and the Pooling1 layer allows for a lower sampling of 1/2 of the input.

The Conv2 layer is also a convolutional layer and consists of eight filters, each of size 3*3.The same padding is also done, and the input to the Pooling2 layer is calculated using the Relu activation function. The Pooling2 layer is also the largest pooling layer, consisting of two 2*2 filters, and the Pooling2 layer can be sampled down to 1/2 of the input. Then it was fed into the Conv3 convolutional layer, which consisted of 16 filters, each of size 3*3.The same fill is also performed, and then the input is calculated to the FC1 layer through the Relu activation function.FC1 layer is Fully Connected layer. After several times of convolution and pooling, we will first "flatten" the multidimensional data, that is, compress the data of the input channel into a one-dimensional array, and then connect it with FC1 layer. Add a dropout layer after the full connection layer to prevent overfitting. Finally, connect to the Softmax layer to classify it.

## 4. Results

### *4.1 The effectiveness of different data sets in identifying mental fatigue*

To evaluate the effectiveness of different data sets in identifying mental fatigue, we divided 2400 EEG fragments into three data sets according To the different utilization rates. The first data set is composed of 1200s experimental data from 1-400s,1001-1400s and 2001-2400s.The second data set is composed of experimental data of 1-500s,951-1450s,1901-2400s, and 1500s in total. The third data set consisted of the experimental data of 1-600s,901-1500s,1801-2400s.

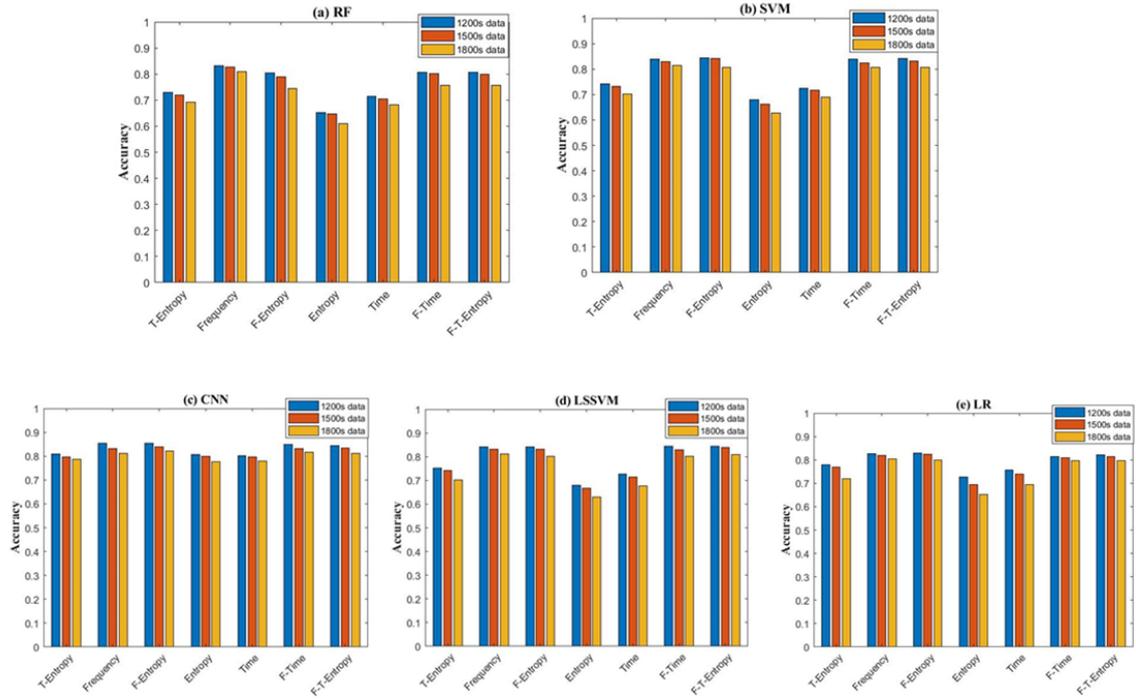

Fig. 5. Comparison of the average precision of the three data sets in RF,SVM,CNN,LSSVM and LR classifiers.

Table 2. the classification accuracy of different data sets in the CNN classifier

| feature | data sets | | |
| --- | --- | --- | --- |
| | 1200s | 1500s | 1800s |
| T_Entropy | **0.8088** | 0.7959 | 0.7865 |
| Frequency | **0.8529** | 08322 | 0.8116 |
| F_Entropy | **0.8534** | 0.8388 | 0.8222 |
| Entropy | **0.8061** | 0.7983 | 0.7766 |
| Time | **0.8026** | 0.7960 | 0.7793 |
| F_Time | **0.8502** | 0.8308 | 0.8156 |
| F_T_Entropy | **0.8444** | 0.8348 | 0.8127 |

According to the experimental results, among the five classifiers of RF,SVM,CNN,LSSVM and LR, the identification accuracy of fatigue level using 1200s data set is higher than that of 1500s and 1800s data sets, and the experiment time of 1200s data set is also the least, so the subsequent studies use 1200s data set.

*4.2 The classification accuracy of different classifiers*

In the assessment of mental fatigue based on EEG signals, it may often be useful to test multiple approaches since there is no uniform classification for all tasks and applications. In this work, we used eight classifiers to test the accuracy. The experimental results show that CNN has the highest classification accuracy.

Least squares support vector machine (SVM) is an improved support vector machine (SVM), it is the traditional support vector machine (SVM) in the inequality constraints into equality constraints, and the error sum of squares (Sum Squares Error) loss function as the experience of the training set loss, thus the problem of solving quadratic programming problem is transformed into solving linear equations, improve the speed of solving problems and convergence precision. Logarithmic probability regression (LR) is a classification model that reduces the prediction range and limits the predicted values to [0,1].

Least Squares Support Vector Machines(LSSVM) and Logistic Regression (LR) were used to process the dataset, and each dataset was divided into three categories: low fatigue, medium fatigue and high fatigue. OVO (One to One, One vs One) classification mode was used to classify the three classifying tasks, and the training set was divided into three sections labeled as low, medium and high respectively. Divided into three dichotomous tasks, the three tasks are {low, medium} respectively;{medium, high};{low, high}.Each task trains a classifier to get three classification results, and the final result is voted: the most predicted category is the final classification result. Least Square Support Vector Machine (LSSVM), the kernel function is linear kernel, and the regularization parameter is set to 2.

SVM is a binary classification model, whose basic model is defined as a linear classifier with the largest interval in the feature space. Its learning strategy is to maximize the interval, which can eventually be transformed into the solution of a convex quadratic programming problem. The kernel function of Support Vector Machines (SVM) is linear kernel and the regularization parameter is set to 1.Random forest is an algorithm that integrates multiple trees through the idea of ensemble learning. Its basic unit is decision tree. In the classification problem, N trees will have N classification results for an input sample. The random forest integrates all the classification voting results and designates the category with the most votes as the final output. Random Forest Classifier (RF) The number of trees in the Forest is 200.K value of K-Nearest Neighbor (KNN) classifier is set to 50.

Table 3. average classification accuracy of the classifiers under different features

| Classifier | feature | | | | | | | |
|---|---|---|---|---|---|---|---|---|
| | T_Entropy | Frequency | F_Entropy | Entropy | Time | F_Time | F_T_Entropy | Mean |
| SVM | 0.7417 | 0.8390 | 0.8454 | 0.6800 | 0.7240 | 0.8382 | 0.8409 | 0.7870 |
| RF | 0.7285 | 0.8309 | 0.8037 | 0.6515 | 0.7139 | 0.8079 | 0.8064 | 0.7633 |
| NB | 0.5567 | 0.6459 | 0.5432 | 0.5224 | 0.5712 | 0.5632 | 0.5528 | 0.5651 |
| CNN | **0.8088** | **0.8529** | **0.8534** | **0.8061** | **0.8026** | **0.8502** | 0.8444 | 0.8312 |
| LSSVM | 0.7509 | 0.8418 | 0.8424 | 0.6787 | 0.7279 | 0.8445 | 0.8454 | 0.7902 |
| LR | 0.7782 | 0.8257 | 0.8292 | 0.7281 | 0.7561 | 0.8152 | 0.8209 | 0.7933 |
| KNN | 0.6302 | 0.7011 | 0.6646 | 0.6003 | 0.6246 | 0.6580 | 0.6571 | 0.6480 |
| DT | 0.5892 | 0.7008 | 0.6443 | 0.5379 | 0.5768 | 0.6629 | 0.6494 | 0.6230 |

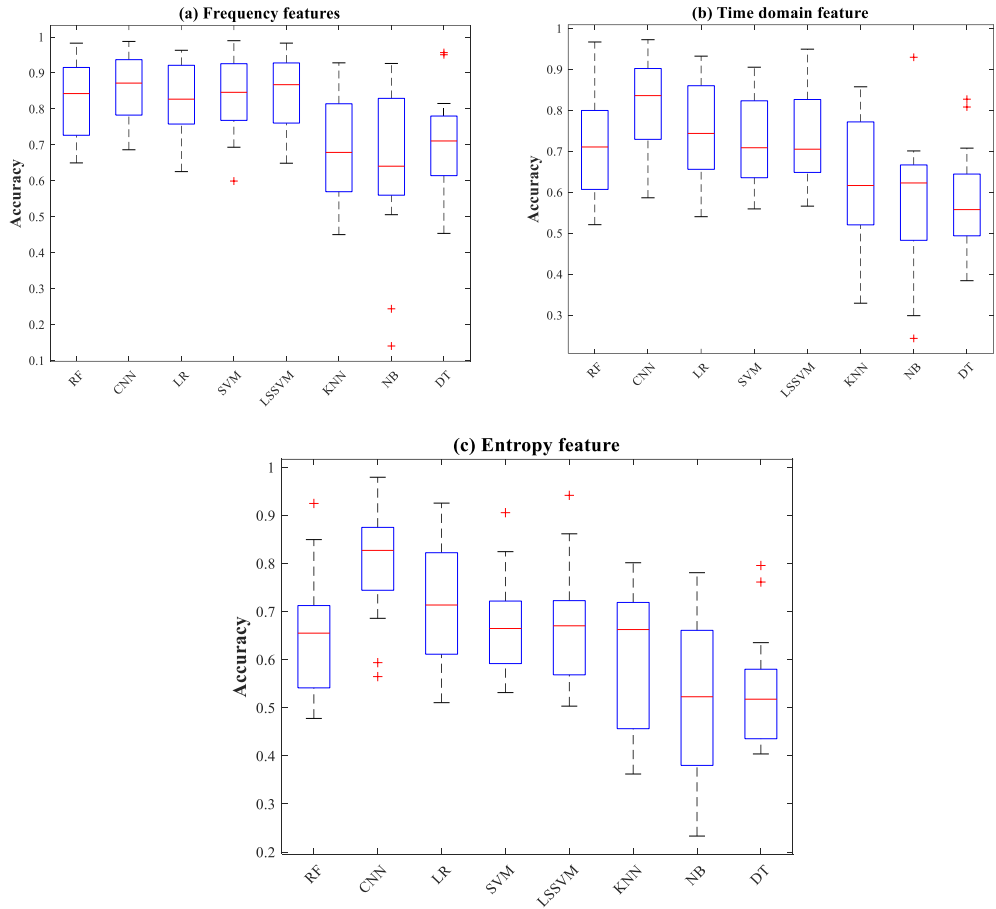

Fig. 6. (a) ~ (c) show the accuracy comparison of 8 classifiers under the same feature

The table fully shows the results of the comparison of different classifiers. In general, the average accuracy of 8 classifiers based on 7 features and 15 subjects is ranked as CNN/ LR/ LSSVM/ SVM/ RF/KNN/DT/NB from largest to smallest. The figure shows the accuracy of each classifier in the same category of features. The figure shows the classification results of each classifier under different feature combinations. The results show that CNN has the best classification effect.

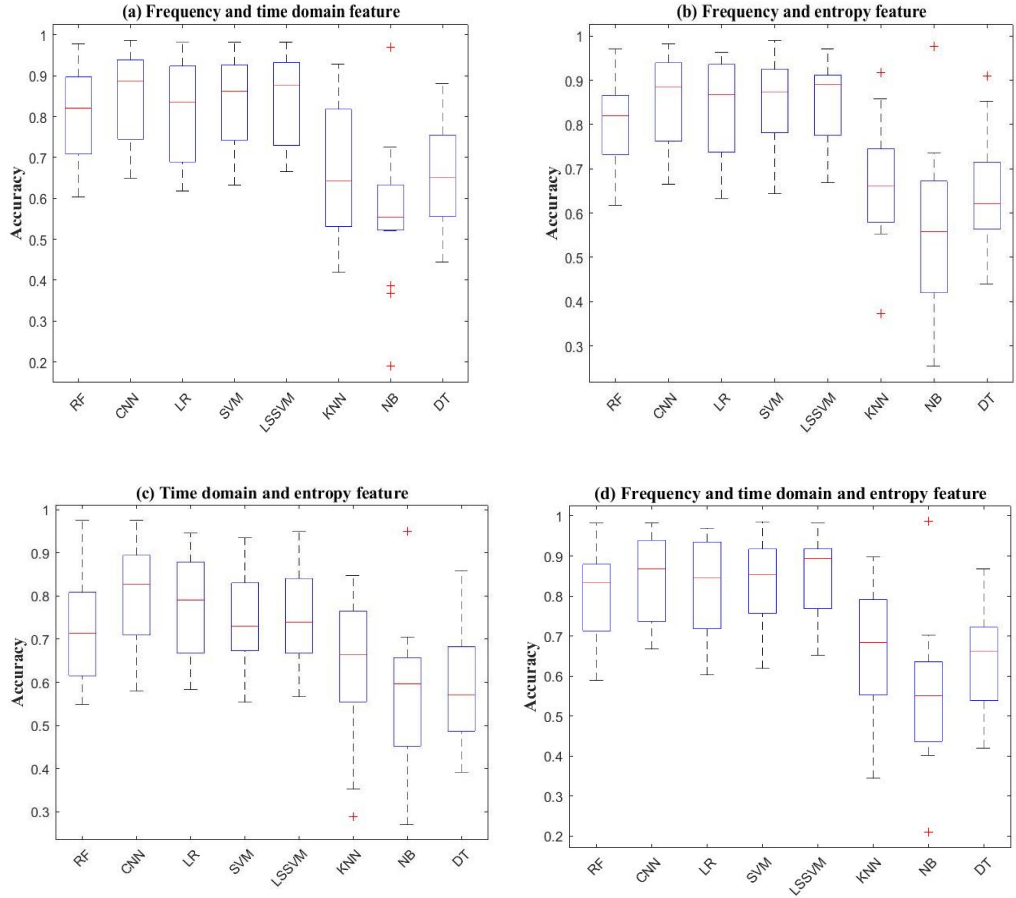

Fig. 7.　(a) ~ (d) show the accuracy comparison of the next 8 classifiers under different feature combinations

## 4.3 Validity of different characteristics in identifying mental fatigue

EEG signals can be regarded as a group of time series. The time domain of time series refers to the relationship between mathematical functions or physical signals and time. The analysis of EEG signal from the frequency domain, time domain, complexity analysis of three aspects. The method used in frequency domain analysis in this paper is power spectrum estimation, whose significance is to transform the EEG wave whose amplitude changes with time into the EEG power changes with frequency spectrum, so that the distribution and transformation of EEG rhythm can be observed intuitively. Time domain analysis methods use the following five: zero crossing analysis, analysis of variance, peak detection, skewness analysis, mean analysis. Complexity analysis is based on Shannon entropy and spectral entropy. Information entropy mainly reflects the uncertainty of information. One of its important functions is to provide a certain basis for judgment when making decisions.

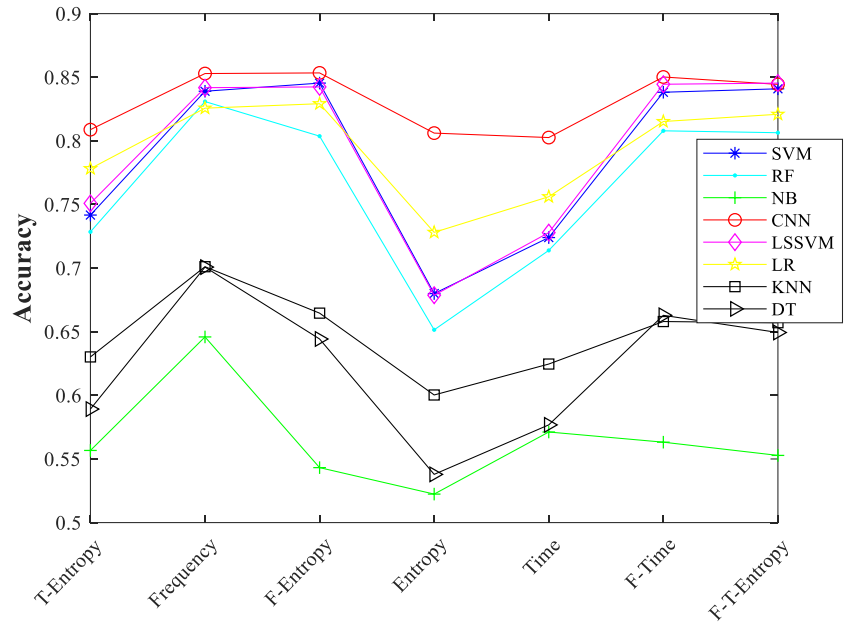

Fig. 8. Comparison of effectiveness of 7 different features in identifying mental fatigue among 8 classifiers.

In order to evaluate the effects of different features and feature combinations on performance, seven feature methods and eight classifiers were compared. In this experiment, EEG signals are analyzed using frequency domain characteristics, time domain characteristics and entropy characteristics. There is also a combination of them: Frequency and time domain feature, Frequency and entropy feature, entropy and time domain feature, Frequency and time domain and entropy feature. The figure shows the average accuracy of the classification accuracy obtained from the seven feature methods using eight classifiers based on EEG signals from 14 channels of 15 subjects. As can be seen from the results in the figure, the combination of Frequency and entropy feature and CNN has the highest classification accuracy, which is 85.34%.The combination of entropy feature and NB has the worst classification accuracy, which is 52.24%.

*4.4 Cross validation*

In the table, we present the classification accuracy obtained by 10-fold cross-validation of a 1200S dataset of 15 subjects. That is, 1200 EEG data from each subject were randomly divided into 10 subsets (each of which had 120 instances). For each iteration, nine subsets are used to train the classifier and the remaining subsets are used for testing. After all subsets have been tested, the average accuracy is calculated and displayed in the table. It can be seen from Tables 3 and 4 that CNN achieves the highest performance with the combination of the features T_entropy, Frequency, F_entropy, Entropy, Time, F_time, and LSSVM achieves the highest accuracy with the combination of the features F_T_entropy. The results show that CNN is superior to the other seven classifiers and F_Entropy is superior to the other six features.

Table 4. Precision of 10-fold cross validation for 15 subjects under 7 features and feature combinations

| Classifier | feature | | | | | | |
|---|---|---|---|---|---|---|---|
| | T_Entropy | Frequency | F_Entropy | Entropy | Time | F_Time | F_T_Entropy |
| A | 0.6440 | 0.6988 | 0.6650 | 0.6861 | 0.5910 | 0.6498 | 0.6908 |
| B | 0.8552 | 0.8718 | 0.8787 | 0.8396 | 0.8539 | 0.8617 | 0.8676 |
| C | 0.9037 | 0.9602 | 0.9616 | 0.9334 | 0.9200 | 0.9480 | 0.9548 |
| D | 0.9617 | 0.9751 | 0.9748 | 0.9462 | 0.9489 | 0.9832 | 0.9819 |
| E | 0.9755 | 0.9877 | 0.9824 | 0.9795 | 0.9725 | 0.9867 | 0.9822 |
| F | 0.7818 | 0.8467 | 0.8463 | 0.7917 | 0.7909 | 0.8284 | 0.8203 |
| G | 0.8275 | 0.8857 | 0.9034 | 0.8622 | 0.8360 | 0.8898 | 0.8843 |
| H | 0.8150 | 0.8840 | 0.8965 | 0.8522 | 0.8346 | 0.8861 | 0.8475 |
| I | 0.8650 | 0.8709 | 0.8993 | 0.8272 | 0.8472 | 0.9098 | 0.9046 |
| J | 0.5807 | 0.6864 | 0.6722 | 0.5939 | 0.5866 | 0.6774 | 0.6675 |
| K | 0.8586 | 0.8848 | 0.8839 | 0.8273 | 0.8416 | 0.8987 | 0.8813 |
| M | 0.6232 | 0.7706 | 0.7469 | 0.5649 | 0.5966 | 0.7172 | 0.7082 |
| N | 0.9411 | 0.9542 | 0.9511 | 0.8795 | 0.9182 | 0.9600 | 0.9508 |
| P | 0.8143 | 0.8192 | 0.8086 | 0.7721 | 0.7926 | 0.8531 | 0.8149 |
| Q | 0.6852 | 0.6972 | 0.7302 | 0.7352 | 0.7087 | 0.7038 | 0.7099 |
| Mean | 0.8088 | 0.8529 | 0.8534 | 0.8061 | 0.8026 | 0.8502 | 0.8444 |

## 4.5 Performance indicators

In order to evaluate the classification effect of each classifier, the well-known performance indicators include accuracy (ACC), sensitivity (SEN), specificity (SPE), accuracy (PRE), and negative predictive value(NPV) described as follows:

$$\text{Psen} = \frac{TP}{TP+FN}, \text{Spe} = \frac{TN}{TN+FP}, \text{Acc} = \frac{TP+TN}{TP+TN+FP+FN},$$

$$\text{Pre} = \frac{TP}{TP+FP}, \text{Npv} = \frac{TN}{TN+FN}$$

Table 5 The confusion matrix of triclassification problems

| Confusion matrix | | Predict | | |
|---|---|---|---|---|
| | | 1 | 2 | 3 |
| Real | 1 | a | b | c |
| | 2 | d | e | f |
| | 3 | g | h | i |

For the three-way classification problem, as shown in the table, Tn =a refers to the number of data inputs in the normal state, and Fn =b+c refers to the number of data inputs classified as the normal state in the fatigue state. FP= d + g refers to the number of data inputs classified as fatigue state in the normal state.TP= e + f + h + i represents the number of data inputs correctly classified as fatigue in the fatigue state.

Table 6. The precision indexes of 7 features and feature combinations in CNN classifier

| Feature | $P_{acc}$ | $P_{npv}$ | $P_{pre}$ | $P_{sen}$ | $P_{spe}$ |
|---|---|---|---|---|---|
| T_Entropy | 0.8088 | 0.9214 | 0.7983 | 0.8339 | 0.7541 |
| Frequency | 0.8529 | 0.9361 | 0.8402 | 0.8654 | 0.8018 |
| F_Entropy | 0.8534 | 0.9399 | 0.8453 | 0.8751 | 0.8090 |
| Entropy | 0.8061 | 0.9091 | 0.7836 | 0.8149 | 0.7305 |
| Time | 0.8026 | 0.9158 | 0.7907 | 0.8252 | 0.7450 |
| F_Time | 0.8502 | 0.9397 | 0.8416 | 0.8723 | 0.8028 |
| F_T_Entropy | 0.8444 | 0.9405 | 0.8432 | 0.8748 | 0.8106 |

The table summarizes the five performance indicators of CNN obtained according to the confusion matrix under seven characteristics, namely, PACC, PNPV, PPRE, PSEN and PSPE.PNPV represents the negative predictive value. Among the samples predicted by the model to be negative, the truly negative samples account for the proportion. The combination of CNN and F_T_Entropy feature has the highest PNPV value, which is 0.9405.Pac represents the accuracy of the model, i.e., the number of correct model predictions/the total number of samples. PPRE represents the positive predictive value, and the proportion of the samples that are truly positive in the positive category predicted by the model. PSEN represents the true class rate, the ratio of the number of samples predicted to be positive in the model to the total number of positive samples. According to the table, under the combination of CNN and F_Entropy feature, PACC, PPRE and PSEN have the highest values, which are 0.8534, 0.8453 and 0.8751, respectively.

## 5. Discussion

In this study, we collected a new database on language understanding tasks and explored the effectiveness of data utilization on classification accuracy. It was found that in the three data sets of 1200s, 1500s and 1800s, the classification accuracy ranges from large to small from 1200s/1500s/1800s, that is, the lower the data utilization is, the better the classification accuracy is. In order to explore the classifier suitable for this study, we used eight classifiers such as CNN,LSSVM and LR to test the accuracy. The experimental results show that CNN has the highest classification accuracy. For CNN's input, Majid et al extracted the Theta, Alpha Beta, and Gamma bands from each segment sample and formed a color RGB image with three channels. Then the generated image is imported into the convolutional neural network for feature extraction and classification [39].In the CNN model, we introduce a new input form, which combines the time-domain, frequency-domain and entropy features extracted from EEG signals and converts them into 3D EEG topographic map as the input of CNN.

In order to evaluate the effectiveness of the feature for mental fatigue classification task, Li et al. collected EEG data of 16 channels and extracted the features of three bands (θ, α, and β), and the results showed that the model using this feature was effective in evaluating mental fatigue of drivers [16].Luo et al. extracted entropy features and classified them, thus achieving a good accuracy in fatigue detection [40].According to the survey, frequency domain features, time domain features and entropy features have good effects on mental fatigue classification model [41]-[42].In this paper, we use seven features and feature combinations and eight classifiers to explore the effectiveness of features. The results show that the combined classification accuracy of CNN and Frequency and Entropy feature is the highest, which is 85.34%.

The limitations and future work of this study include the following two aspects.(1) We only use three convolutional layers and two pooling layers in the CNN classifier. In the future work, we will enrich the diversity of CNN structure through deeper structure to extract more significant features and obtain better classification accuracy. (2) For EEG feature extraction, this study uses frequency domain feature, time domain feature and entropy feature. In future studies, we can try the differential entropy feature and the differential entropy feature after wavelet transform to evaluate the effectiveness of different features for mental fatigue classification.